\documentclass[11pt,a4paper]{article}
\usepackage[hyperref]{emnlp2020}
\usepackage{times}
\usepackage{latexsym}

\usepackage{amsmath}
\usepackage{amsthm}
\DeclareMathOperator*{\argmax}{arg\,max}

\usepackage{enumitem}
\usepackage{microtype}

\aclfinalcopy % Uncomment this line for the final submission
\title{Why Neural Machine Translation Prefers Empty Outputs}

\author{{\em Xing Shi, Yijun Xiao, and Kevin Knight}  \\
 \\
 \\
 \\
{\tt \{kevinknight\}@didiglobal.com}}

\author{Xing Shi \\ DiDi Labs \\  4640 Admiralty Way \\ Marina del Rey
\And Yijun Xiao\thanks{\ \ research performed while at DiDi Labs} \\  Department of Computer Science \\ UC Santa Barbara
\And Kevin Knight \\ DiDi Labs \\  4640 Admiralty Way \\ Marina del Rey}

\date{}

\begin{document}

\maketitle

\begin{abstract}

We investigate why neural machine translation (NMT) systems assign high probability to empty translations. We find two explanations. First, label smoothing makes correct-length translations less confident, making it easier for the empty translation to outscore them. Second, NMT systems use the same, high-frequency EoS word type to end all target sentences, regardless of length. This creates an implicit smoothing that increases the relative probability zero-length translations.  Using different EoS types in target sentences of different lengths exposes this implicit smoothing.
\end{abstract}

\section{Introduction}

We typically train neural machine translation (NMT) systems on human-translated parallel texts, then ask them to decode previously-unseen source sentences.  Trained parameter values induce a distribution $P(y | x)$ over all pairs of source/target strings ($x, y$).  Given a new source string $x$, the NMT decoder searches for the best target string $y$:

\begin{align}
    \hat{y} = \argmax_y P(y \mid x)
\end{align}

This optimization is unsolvable for general recurrent neural networks \cite{Chen2018RecurrentNN}, while \citet{byrne2019} present an exact optimization search algorithm for consistent NMT models.  

\begin{table}
\centering
\begin{tabular}{c r r}
\hline
Dataset & Train & Test\\ 
\hline
English / German & 4,508,785 & 200\\ 
Chinese / Japanese & 19,201,050 & 200\\ 
\hline
\end{tabular}

\vspace*{0.1in}

\begin{tabular}{l | r r } \hline
Language & \multicolumn{2}{|c}{Decoder beam = 512} \\
pair & Length ratio & Empty ratio \\ \hline
DE$\rightarrow$EN & 0.257 & 0.525 \\%& 30.33\\
EN$\rightarrow$DE & 0.305 & 0.520 \\ %& 25.09\\ 
JA$\rightarrow$ZH & 0.773 & 0.030 \\% & 32.60\\ 
ZH$\rightarrow$JA & 0.777 & 0.010 \\%& 33.79\\ 
\hline
\end{tabular}
\caption{For four language pairs, we decode 200 test sentences each, using NMT systems trained with the Tensor2Tensor toolkit \cite{tensor2tensor}.  Length ratio shows MT-output tokens divided by reference tokens.  Empty ratio shows the percentage of source sentences that result in zero-length target outputs.  A large decoder beam size (512) exposes the well-known fact that NMT learns to assign high probabilities to short/empty translations.}
\label{table:problem}
\end{table}

In practice, we build up a target string $y$ using a left-to-right, word-by-word greedy strategy.  All target sentences end with the pseudo-word EoS, in both train and test data.  When the greedy search selects EoS, the translation ends.  

We can easily find higher-probability strings $y$ with beam search \cite{Sutskever2014SequenceTS}.  However, when we use a large beam, the higher-probability strings turn out to be worse translations, as judged by both Bleu \cite{Papineni2001BleuAM} and human evaluators.  In fact, the highest-probability string is often very short, or even empty (i.e., a one-word translation ``EoS'').  

We therefore typically revert back to a small beam size, hoping for a good translation despite a worse $P(y | x)$.  When this happens, we have a ``fortuitous'' search error \cite{fortuitous}.  As NMT system architectures have moved from LSTM recurrent neural networks \cite{lstm} to self-attention Transformer models \cite{transformer}, the empty translation problem has lessened a bit, but is still very present \cite{byrne2019}. 

Table~\ref{table:problem} shows the behavior of four transformer-based NMT models trained on German-English and Chinese-Japanese parallel data, using decoder beam size~512. The {\em length ratio} is the token ratio of the generated translations compared to reference translations. The {\em empty ratio} is the percentage of empty translations (only the EoS token) over the source sentences. We see that around half the translations on German-English models are empty.

Our central question is: {\em why are empty translations preferred?}  Our training data does not contain any source strings translated to empty strings, so why does NMT learn to assign high probability to empty translations? Our findings are: 

\begin{itemize}
\item The reference translation generally outscores the empty translation, but the popular label smoothing technique intentionally decreases the probability of the reference translation at each token, by about 5\% absolute.  The cumulative effect is to make the empty translation more probable than the reference.

\item The popularity of EoS in the target corpus seems to increase its overall attractiveness as a generation candidate at any target length.  If we use different EoS types for different target lengths, we find that empty translations are no longer preferred.
\end{itemize}

\section{Related Work}

A short translation has only a few factors in its probability score, e.g., P(\text{EoS} $\mid x, y_1, y_2$), whereas a translation $y$ of appropriate length $m$ will have many more factors:

\begin{align}
    P(y \mid x) = & P(y_1 \mid x) \cdot \bigg[ \prod_{i=2}^m P(y_i \mid x, y_1 \ldots y_{i-1}) \bigg] \nonumber\\ 
    & \cdot P(\text{EoS} \mid x, y_1 \ldots y_m)
\end{align}

Therefore, we might imagine that a short translation naturally has a very high probability score that must be overcome by other means.  In fact, these other means indeed exist.  Statistical MT (SMT) researchers \cite{Brown1990ASA,Och1999ImprovedAM} developed several mechanisms for regulating target length: 

\begin{itemize}
    \item IBM Model 1.  A conditional length parameter $\epsilon(m \mid l)$ scores target length $m$ given source length $l$.  After training on clean parallel corpora, an entry like $\epsilon(0 \mid 21)$ would be zero.
    \item IBM Model 3.  A fertility model parameter $\phi(n \mid x_i)$ forces a decision, for each source word, about how many target words to generate.  Generating an empty translation would require $l$ separate choices of the form $\phi(0 \mid x_i)$, most of which will be low after training.
    \item Coverage vector.  In phrase-based translation, the decoder only stops when every phrase in a phrase-chunked source sentence produces some non-empty target phrase.
\end{itemize}

\noindent
While these mechanisms strongly mitigate against short target translations, SMT designers found that a word-bonus is still needed.\footnote{``\ldots the product P(e) P(f$|$e) is too small for long English strings as compared with short ones. As a result, short English strings are improperly favored over longer English strings.'' \cite{Brown1990ASA}}

Neural MT (NMT) offered the welcome prospect of automatically-learned length regulation. \citet{Shi2016WhyNT} show how an individual hidden unit learns to count source words during encoding (by decrementing its activation) and count target words during decoding (by incrementing its activation). During greedy decoding, the probability P(EoS) spikes around the time the unit's activation returns to zero.

We therefore expect sequence-to-sequence models to rule out short translations when trained with maximum likelihood.  That is, the counting units should hold down P(EoS) until an appropriate time.  If the network assigns a high probability to a very short output, then further training epochs should move probability away from that output and towards the reference translation.

\paragraph{Documenting the problem.}  Many papers describe the extent of the length problem. \newcite{byrne2019} use exact decoding to find that up to half of the highest-scoring translations in a WMT~2015 task are empty translations.
\newcite{yang2018breaking} show that shorter translations have higher model scores and note that these will eventually be found by search with a large enough beam.

\paragraph{Proposing fixes.}  Most prior work focuses on proposing and evaluating practical fixes for the length problem. For example, length normalization divides the model score by the length of the output \cite{jean2015montreal,murray2018correcting}. The NMT system of \citet{wu2016google} implements a more complicated correction by dividing the model score by a parameterized function of the length. \citet{he2016improved} adds a constant word bonus to promote longer candidates, borrowing an idea from SMT.  \citet{yang2018breaking} propose a hyperparameter-free re-scoring method.  However, Bleu rankings in MT shared tasks are still strongly affected by human designers guessing which length-bonus parameter value will work best on a hidden test set.

\paragraph{Explaining the problem.} Other prior work suggests root causes for the length problem.  \citet{sountsov2016length} claim that the maximum-likelihood training objective underestimates the margin separating long sequences from short sequences. Therefore the loss function is not aligned with the ideal goal of making the global margin between the reference and incorrectly short candidates positive. \citet{murray2018correcting} attribute the length problem to label bias and local normalization.  \citet{ott2018analyzing} claim that degradation with larger beam size is, at least in part, due to the intrinsic and extrinsic uncertainties of the task. They show that introducing higher uncertainty to the data causes further degradation.  \citet{cohen2019empirical} use the concept of search {\em discrepancies} (deviations from greedy choices), observing that earlier positions have larger discrepancies with larger beam size.  Some authors place the blame on the token-level maximum likelihood objective.  Indeed, researchers find that replacing maximum likelihood with maximum expected Bleu causes the length problem to disappear \cite{maxbleu}.

However, none of these papers directly solves the mystery outlined in Section~1, i.e., why doesn't maximimum likelihood training already shift probability away from emtpy output candidates, toward appropriate-length ones?

\section{Data}

In this work, we train NMT models on the data in Table~\ref{table:problem}:

\begin{itemize}
    \item English-German: We use the standard WMT 2014 English-German dataset together with news-commentary-v12 data from WMT 2017 as the training data. We remove all the pairs whose source or target sentence length is 0. We use newstest2014 as testset.  
    \item Chinese-Japanese: The training data consists of both existing parallel data and web-crawled parallel data. The test data include daily expressions, news, and dialogues.
\end{itemize}

We use byte-pair encoding \cite{sennrich-etal-2016-neural} with 30K merge operations to generate the vocabulary.  All NMT models are trained using the Tensor2Tensor toolkit\footnote{https://github.com/tensorflow/tensor2tensor} \cite{tensor2tensor} with transformer-base hyper-parameters. The label smoothing is 0.1 and the mini-batch size is 4096 tokens. All the models are trained on a single Nvidia P40 GPU with 24Gb device memory, for 500 thousand steps (3 days).

Since our aim to is understand why NMT learns to assign high probability to empty translations, and not to improve state-of-the-art Bleu scores, we do not carry out detailed, comparative Bleu evaluations.  To support the soundness of our implementation, we report baseline Bleu scores of 30.33 for DE$\rightarrow$EN (standard 3003-sentence test set) and 25.09 for EN$\rightarrow$DE.

\section{When Empty Translation Are Preferred}

\begin{table*}
\centering
\small
\begin{tabular}{l l|c c c c c}
\hline
Language & Label & \multicolumn{4}{c}{Decoder beam = 512 } \\
pair &   smoothing & Length ratio & Empty ratio & $\min \log P([j \in V]|x)$ & $\log P([EoS]|x)$ & $\frac{\log P(y|x)}{|y|}$ \\ \hline
DE$\rightarrow$EN & 0.0 & 0.35 & 0.430 & -27.50 & -8.94 & -0.40 \\ 
 & 0.1 & 0.26 & 0.525 & -17.67 & -9.41 & -0.51  \\ \hline
EN$\rightarrow$DE & 0.0 & 0.36 & 0.425 & -23.54 & -9.47& -0.38 \\
 & 0.1 & 0.31 & 0.525 & -17.52 & -9.65 & -0.48  \\ \hline
JA$\rightarrow$ZH & 0.0 & 0.84 & 0.020 & -24.83 & -12.78 & -0.61 \\
 & 0.1 & 0.77 & 0.030 & -17.74 & -11.78 & -0.72  \\ \hline
ZH$\rightarrow$JA & 0.0 & 0.89 & 0.005 & -24.45 & -12.95 & -0.47 \\ 
 & 0.1 & 0.78 & 0.010 & -18.65 & -11.97 & -0.57  \\ \hline
\end{tabular}
\caption{Behavior of 4 NMT models with and without label smoothing. $\min \log P([j \in V]|x)$ is the minimum log probability of any word in the vocabulary when decoding the first token. $\log P([EoS]|x)$ is the log probability of EoS word in first target position, given the source sentence $x$. $\frac{\log P(y|x)}{|y|}$ is the log probability of the reference target sentence $y$ given source sentence $x$, divided by the target sentence length, i.e. the token-level average log probability. Bleu scores are calculated on full test set.} 
\label{table:ls}
\end{table*}

Given a source sentence $x = [x_1, x_2, ... , x_m]$ and a target sentence $y = [y_1, y_2, ..., y_l, EoS]$, a trained NMT model will provide the conditional probability $P(y|x)$. Let $Y_k$ denote the set of sentences that will be explored during the beam search with beam size k and whose length is larger than 0. When we also score the empty translation, we often find it is preferred:

\begin{align}
    \forall_{y \in Y_k} \ P([EoS]|x) > P(y|x) \label{equation:reason}
\end{align}

So the empty translation may be preferred for two reasons; either (1) $P([EOS]|x)$ is not small enough, or (2) $P(y \in Y_k|x)$ is not large enough.

\section{Label Smoothing}

Label smoothing is a popular technique regularizing the output distributions to alleviate over-confident predictions. The cross-entropy loss with label smoothing is: 
\begin{align}
    L_\epsilon = - (1-\epsilon) \log p(i) - \sum_{j\neq i \in V} \frac{\epsilon}{|V|-1} \log p(j)
\end{align}
where $V$ is the set of vocabulary, $i$ is the correct token and $\epsilon$ is a small positive hyper-parameter for label smoothing. 

With label smoothing, $p(i)$ is optimized toward $1-\epsilon$ instead of $1$, and $p(j\neq i)$ will be optimized toward $\frac{\epsilon}{|V| - 1}$ instead of $0$. Thus, in theory, label smoothing will increase the probability of incorrect tokens, including the EoS token at the first decoding position, i.e., $P([EoS]|x)$. It will also decrease the probability of correct tokens, so that $P(y\in Y_k|x)$ will decrease. 

Table~\ref{table:ls} shows the behavior of four NMT models trained with and without label smoothing. For all four translation directions, label smoothing causes a higher empty ratio:

\begin{itemize}
    \item Label smoothing increases the lower bound of the log probability of any word when decoding the first output token, i.e., $\min \log P([j \in V]|x)$. However, $\log P([EoS]|x)$ doesn't always increase. For DE-EN and EN-DE, $\log P([EoS]|x)$ decreases when we turn on label smoothing. 
    \item The second-rightmost column of Table~\ref{table:ls}, $\frac{\log P(y|x)}{|y|}$, shows the token-level average log probability of the reference target sentence. This value decreases when we turn on label smoothing (by about 5\% absolute), indicating that label smoothing decreases $P(y \in Y_k|x)$. 
\end{itemize}

To make this concrete, suppose we are translating a German sentence $x$, and we have a 20-word candidate English translation $y$. Without label smoothing, $\log P(y|x)$ will be around $-0.40 * 20 = -8.0$, which is higher than the log probability of empty translation $-8.94$. With label smoothing, $log P(y|x)$ will be roughly $-0.51 * 20 = -10.2$, which will be lower than $\log P([EoS]|x)$, which is $-9.41$. Thus, the empty translation will be preferred.

In summary, label smoothing decreases the probability of normal length translation, but it does not necessarily increase the probability of the empty translation.  It obtains more empty translations mainly by affecting the right hand side of Equation~\ref{equation:reason}, rather than both sides.  
\section{Single-EoS Smoothing Effect}

There are still two mystery gaps in Table~\ref{table:ls}:

\begin{enumerate}
    \item Why is there a large gap of $\log P([EoS]|x)$ between English-German models and Chinese-Japanese models? The vocabulary sizes of all four directions are quite similar. 
    \item For all models, why is there a larger gap between $\log P([EoS]|x)$ (EoS in the first position) and $\min \log P([j\in V]|x)$ (the least likely word in the first position)? Since there are no pairs in training data with target length zero, EoS should be among the ``worst'' words in the first position.
\end{enumerate} 

\begin{table}
\centering
\small
\begin{tabular}{c c c c c}
\hline
 & Perplexity & Empty & \\
Model &  of $Q(l|m)$ &  ratio & $log P([EoS]|x)$\\ 
\hline
JA-ZH & 12.0  & 0.020 & -12.77\\ 
EN-DE & 27.4  & 0.425 & -9.47\\ 
EN-DE-75 & 12.4  & 0.200 & -11.43\\ 
EN-DE-50 & 6.9  & 0.010 & -13.26\\ 
\hline
\end{tabular}
\caption{The perplexity of $Q(l|m)$, empty ratio, and $\log P([Eos]|x)$ of four different training sets. High perplexity indicates high uncertainty for a distribution. The models here are trained without label smoothing.} 
\label{table:ppx}
\end{table}

\begin{table*}
\centering
\small
\begin{tabular}{c c|c c c c}
\hline
 & &\multicolumn{4}{|c}{Decoder beam = 512}  \\
Model & EoS  & Length ratio &  Empty ratio & $\min \log P([j \in V]|x)$ & $\log P([EoS]|x)$ \\ 
\hline
DE-EN & Single & 0.26 & 0.525 & -17.67 & -9.41 \\%& 30.33\\ 
 & Multi & 0.74 & 0.025 & -30.80 & -30.7 \\%& 30.33\\ 
\hline
EN-DE & Single & 0.31 & 0.525 & -17.52 & -9.65 \\%& 25.09\\ 
 & Multi& 0.86 & 0.020 & -20.48 & -20.40 \\%& 25.03\\ 
\hline
JA-ZH & Single &0.77 & 0.030 & -17.74 & -11.78 \\%& 32.60\\ 
 & Multi & 0.80 & 0.000 & -22.20 & -22.02 \\%& 32.53\\ 
\hline
ZH-JA & Single & 0.78 & 0.010 & -18.65 & -11.97 \\%& 33.79\\ 
 & Multi & 0.87 & 0.005 & -21.90 & -21.54 \\%& 33.99\\ 
\hline
\end{tabular}
\caption{Behavior of four NMT models trained with Single EoS type and Multiple EoS types, respectively. All models are trained with label smoothing 0.1.} 
\label{table:multieos}
\end{table*}

We find that these two gaps can be explained by an implicit smoothing effect caused by the basic design of the EoS, i.e., that all target sentences of different lengths will end with the {\em same} EoS type. To generate a target sentence with proper length, an NMT model needs to make the probability of EoS at step t, $P(EoS_{t})$, close to 1 at proposed positions and close to 0 at other positions. We have:   
\begin{align}
    P(EoS_t) \propto \exp(h_t * e_{EoS})
\end{align}
where $h_t$ is the hidden vector at step $t$ and $e_{EoS}$ is the embedding of EoS token. While $h_t$ varies at different time steps, $e_{EoS}$ remains the same. Thus $P(EoS_0)$ will be smoothed even if EoS never appears in the first position in training data. We refer this implicit smoothing effect as the \textit{single-EoS smoothing effect}.

Given a source sentence with length $m$, there is a natural distribution of the target length $l$, denoted as $Q(l|m)$. The different $\log P([EoS]|x)$ in different language pairs is caused by different uncertainties in $Q(l|m)$. Because of the single-EoS smoothing effect, the $Q(l|m)$ with higher uncertainty will assign more probability mass to unseen target lengths, for example, zero length. Thus the $\log P([EoS]|x)$ will be higher. 

Table~\ref{table:ppx} shows the perplexity of $Q(l|m)$, empty ratio, and $\log P([Eos]|x)$ for four training sets: JA-ZH, EN-DE, EN-DE-75 and EN-DE-50. High perplexity indicates high uncertainty. The perplexity of JA-ZH is much smaller than EN-DE, thus $Q(l|m)$ is more certain and there is less smoothing toward unseen target lengths. In those cases, $\log P([EoS]|x)$ is estimated as smaller, resulting in fewer empty translations. We verify this logic by generating a new training set EN-DE-75, in which we remove sentence pairs of length ($m$, $l$) if $l$ is not among the most frequent 75\% of target lengths for source length $m$. We find that EN-DE-$k$ has lower $Q$ perplexity, lower $\log P([EoS]|x)$, and  fewer empty translations.

If we use different EoS types for target sentences with different lengths, we remove this implicit smoothing.  During training, we append each length-$l$ target sentence with token ``[EOS-l]'', instead of just ``[EOS]''. In total, we add 512 new EoS types to the original vocabulary. During training, we do not conduct label smoothing for these EoS tokens. During decoding, we only stop if we meet ``[EOS-l]'' at step $l+1$. 

We refer this set-up as \textit{MultiEoS}. Table~\ref{table:multieos} shows four NTM models trained with the usual Single EoS versus MultiEoS. Under MultiEoS, EoS takes its rightful place among the ``worst'' words for the first target position.  That is, $\log P([EoS]|x)$ is now close to $\min \log P([j\in V]|x)$).  Furthermore, the preference for empty translations virtually disappears.

\section{Conclusion}

We investigate why NMT systems assign high probability to empty translations. We find that label smoothing mainly decreases the probability of normal-length target sentences, and that the single-EoS smoothing effect increases the probability of the empty translation. 

\bibliography{emnlp2020}
\bibliographystyle{acl_natbib}

\end{document}